# A Generative Adversarial Network-based Selective Ensemble Characteristic-to-Expression Synthesis (SE-CTES) Approach and Its Applications in Healthcare

Yuxuan Li[1], Ying Lin[2], Chenang Liu[1]*

[1]The School of Industrial Engineering & Management, Oklahoma State University, Stillwater, OK
[2]Department of Industrial Engineering, University of Houston, Houston, TX
*Corresponding author. chenang.liu@okstate.edu

**Abstract:**

Investigating the causal relationships between characteristics and expressions plays a critical role in healthcare analytics. Effective synthesis for the expressions using given characteristics can make great contributions to health risk management and medical decision-making. For example, predicting the resulting physiological symptoms on a patient from given treatment characteristics is helpful for the disease prevention and personalized treatment strategy design. Therefore, the objective of this study is to effectively synthesize the medical expressions based on given characteristics for healthcare analytics. However, there are two major challenges: (1) the mapping from characteristics to expressions is usually from a relatively low dimension space to a high dimension space, but most of the existing methods such as regression models are not able to effectively handle such mapping; and (2) the relationship between characteristics and expressions may contain not only deterministic patterns, but also stochastic patterns, i.e., not a simple deterministic one-to-one mapping. To address these challenges, this paper proposed a novel selective ensemble characteristic-to-expression synthesis (SE-CTES) approach inspired by the generative adversarial network (GAN). The novelty of the proposed method can be summarized into three aspects: (1) GAN-based architecture for deep neural networks are incorporated to learn the relatively low dimensional mapping to relatively high dimensional mapping containing both deterministic and stochastic patterns; (2) the weights of the two mismatching errors in the applied GAN-based architecture are proposed to be different to reduce the learning bias in the training process; and (3) a selective ensemble learning framework is proposed to further reduce the prediction bias and improve the model robustness, i.e., improve the synthesis stability.

To validate the effectiveness of the proposed approach, extensive numerical simulation studies and a real-world healthcare case study were applied and the results demonstrated that the proposed method is very promising.

# 1 Introduction

Understanding the underlying relationships between characteristics and the resulting expressions could improve the performance of analytics significantly in many fields (Figure 1), e.g., health monitoring, manufacturing, and bioinformatics. For example, the treatments of metabolic disorders, such as the lifestyle interventions and metabolic medications, could directly affect the related physiological symptoms such as the BMI, blood pressure on patients. The treatment or intervention strategies are regarded as characteristics while the resulting physiological symptoms are annotated as expressions. Accurately inferring the resulting physiological symptoms from treatment strategy is critical for treatment effect evaluation and medical decision making [1]. Moreover, an effective and efficient methodology to handle the characteristics to expressions synthesis can also greatly contribute to the investigation of many other fundamental research problems, such as disease progression analysis in health informatics [2], gene expression modeling and prediction in bioinformatics [3, 4], as well as process monitoring and quality prediction in advanced manufacturing [5, 6].

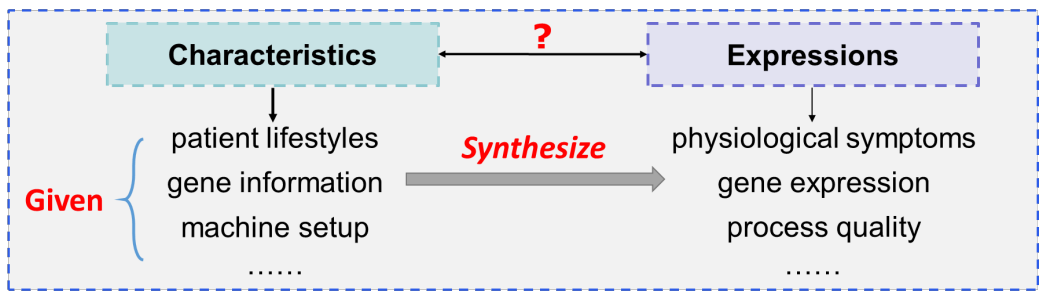

**Figure 1:** *A demonstration of the objective for characteristic-to-expression synthesis*

Characteristics are usually given or can be controlled directly. However, the expressions resulted from a specific setting of characteristics are usually hard to be observed in reality. In the disease treatment strategy design, for example, it is challenging to collect and evaluate the expressions of physiological symptoms under different treatment characteristics on the same patient. Thus, it is commonly existing that the expressions are unknown while the characteristics are given. Under such circumstances, as described in Figure 1, the unknown expressions could be synthesized by the characteristics. Hence, a characteristic-to-expression synthesis approach should be developed to synthesize the expressions based on the given characteristics.

Therefore, the objective of this paper is to develop an effective approach to synthesize the unknown high dimensional expressions based on given characteristics. Mathematically, it can also be formulated as a nonlinear regression problem, where the characteristics and expressions can be treated as the predictors and responses, respectively. Although there are extensive existing regression models, most of them are not capable enough of handling this problem directly due to two key challenges. First, the characteristics usually span in a relatively low dimensional space while expressions are usually high dimensional. For example, in healthcare systems, the characteristics, i.e., the treatment or intervention strategies, may be just a few numerical or categorical variables, but the expressions might be curves or images [7]. However, most of the existing regression models mainly focus on the mapping from high dimensional predictors to low dimensional response variables, which are not appropriate to be directly used for this problem. More importantly, the mapping usually contains both deterministic and stochastic patterns, not pure deterministic patterns. For example, even under the same treatment characteristics, the physiological symptoms will not be unique but contain stochastic patterns, which cannot be ignored [2]. Unfortunately, most of the regression models are designed to capture the deterministic mapping between variables, and thus they may not be able to consider the stochastic patterns among variables. Compared to the existing regression methods, the recent emerging generative adversarial network (GAN) [8]-based approaches, such as GAN-CLS [9], has the ability to learn the joint underlying distribution from the high dimensional data and capture the stochastic patterns. Inspired by GAN-CLS, this work adopts a GAN-based architecture called characteristic-to-expression synthesis (CTES) model. However, due to the application of mismatching pairs in GAN-CLS, the estimation of the underlying data distribution becomes biased. Besides this, the capability of CTES is still limited because of the significant synthesis instability. The instable synthesis process demonstrates that the robustness of the CTES is low. Therefore, in order to reduce the prediction bias and make the synthesis process more stable, several proposed CTES models are ensembled and selected according to the quality of synthesized expressions. Then such architecture is considered as a selective ensemble learning architecture.

Therefore, this paper proposes a novel GAN-based approach termed selective ensemble characteristic-to-expression synthesis (SE-CTES), which is able to synthesize accurate expressions given characteristics. In the proposed method, the contributions mainly consist of three aspects:

(1) The generative adversarial learning architecture for deep neural networks are incorporated to learn the complex mapping from relatively low dimensional characteristics to relatively high dimensional expressions.

(2) The weights of the two mismatching errors in model training are considered differently and further optimized to reduce the learning bias by making the trained data distribution dominate the estimation to the actual data distribution.

(3) A new selective ensemble learning framework is proposed to improve the model robustness, i.e., improve the training stability, and further reduce the prediction bias, by applying a novel inverse validation framework, which inversely correlates synthesized expressions and characteristics to validate the effectiveness of the proposed model.

The rest of this paper is structured as follows. Sec. 2 provides the problem formulation and a brief literature review of the related research. Sec. 3 introduces the proposed research methodology in detail, followed by a simulation study and a real-world case study for the validation of the proposed method in Sec. 4.

Afterwards, Sec. 5 demonstrates an extended simulation study for scalar to matrix synthesis to further demonstrate the capability of the proposed method. Finally, the conclusions are discussed in Sec. 6.

## 2 Research Problem and Literature Review

As discussed in Sec. 1, this paper aims to synthesize expressions using characteristics based on GAN. Thus, this section first formulates the research problem and reviews the related studies as well as their shortcomings (Sec. 2.1). Subsequently, an introduction of GAN-based approaches is presented in Sec. 2.2.

### 2.1 Problem formulation

As discussed in Sec. 1, the objective is to synthesize new expressions based on new characteristics. The existing characteristics are denoted as $\mathbf{x}$ with $m$ variables $[x_1, x_2, ..., x_m]$ and the existing expressions are denoted as $\mathbf{y}$ with $n$ variables $[y_1, y_2, ..., y_n]$ where $n > m$. As shown in Eq. (1), to quantify the relationship between $\mathbf{x}$ and $\mathbf{y}$, a natural idea is to build a model $f: \mathbb{R}^m \to \mathbb{R}^n$ from $\mathbf{x}$ to $\mathbf{y}$. Based on the regression model, it is possible to obtain what expressions the new characteristics, $\mathbf{x}^{\text{new}}$, correspond to.

$$\mathbf{y} = f(\mathbf{x}) \Rightarrow f(\mathbf{x}^{\text{new}}) = ? \tag{1}$$

Subsequently, the problem is how to train $f$ and predict new expression $\hat{\mathbf{y}}^{\text{new}}$ based on a new characteristic $\mathbf{x}^{\text{new}}$. Denote that there are $N$ samples for characteristic $\{\mathbf{x}^{(1)}, \mathbf{x}^{(2)}, ..., \mathbf{x}^{(N)}\}$ and expressions $\{\mathbf{y}^{(1)}, \mathbf{y}^{(2)}, ..., \mathbf{y}^{(N)}\}$. Then the samples could be applied to approximate $f$, i.e., $\hat{f}$. Afterwards, a new expression $\hat{\mathbf{y}}^{\text{new}}$ could be predicted based on $\hat{f}$. Thus, the synthesis can be formulated as Eq. (2),

$$\mathbf{y}^{(i)} = \hat{f}(\mathbf{x}^{(i)}), 1 \leq i \leq N \Rightarrow \hat{\mathbf{y}}^{\text{new}} = \hat{f}(\mathbf{x}^{\text{new}}) \tag{2}$$

Notably, $f$ should have two important properties. First, according to the dimensions of characteristics and expressions, $f$ is a mapping from a relatively low dimension space to a high dimension space. Besides, since one specific characteristic may lead to different expressions by $f$ in practice, inherent stochastic patterns should also be considered in the mapping in addition to the deterministic patterns. In order to approximate $f$, a natural idea is to apply regression approaches. Fernández-Delgado *et al.* [10] provided a detailed literature review for both. However, most of the regression models could not consider both properties simultaneously. For instance, many conventional statistical regressions and discriminative approaches, such as logistic regression [10], ridge regression [11], Bayesian models [12], $k$ nearest neighbors (kNN) [13], support vector machine (SVM) [14] and decision tree [15], could not handle the relatively low dimension to high dimension mapping because the responses of these models are usually low dimensional. Although some regression approaches could handle the low-dimensional-to-high-dimensional mapping, such as partial least square [16], general regression neural network (GRNN) [17] and extreme learning machine (ELM) [18], they are designed to capture the deterministic mapping between variables. Thus, they may not be able to consider the stochastic patterns among variables.

To take the stochastic patterns into account, the popular generative models, such as variational autoencoder (VAE) [19, 20], flow-based model [21, 22], and generative adversarial network (GAN) [8] have this capability. Specifically, for the VAE models, based on the encoding to latent vectors and decoding latent vectors, the data could be reconstructed. Similarly, the flow-based model could also reconstruct the data based on the process of flow and inverse incorporating latent vectors. However, according to the model design of both VAE models and flow-based models, latent vectors are usually low dimensional representations compared with the into data for reconstruction. Hence, these methods may not be suitable to learn the mapping from relatively low dimensional characteristics to relatively high dimensional expressions accurately. Under such circumstances, generative adversarial network (GAN) [8]-based frameworks, become the promising solutions since GAN-based applications in other synthesis tasks, such as the text-to-image synthesis [9], has already demonstrated its higher potential than other existing methods. The literatures related to GAN-based approaches are reviewed in Sec. 2.2.

## 2.2 Generative adversarial network (GAN) based approach

GAN is first introduced by Goodfellow [8] in 2014. It involves two networks: generator $G$ and discriminator $D$. $G$ is to synthesize artificial samples while $D$ is to distinguish whether the sent samples are from actual data or artificial data. These two networks compete with each other through a mini-max game shown in Eq.

(3), where **x** are the samples from actual data distribution $P_{\text{data}}$ and $G(\mathbf{z})$ are the samples from artificial data distribution $P_{G(\mathbf{z})}$.

$$\min_G \max_D V(D, G) = \mathbb{E}_{\mathbf{x} \sim P_{\text{data}}}[\log(D(\mathbf{x}))] + \mathbb{E}_{G(\mathbf{z}) \sim P_{G(\mathbf{z})}}[\log(1 - D(G(\mathbf{z})))] \tag{3}$$

GAN is initially developed for image augmentation [8]. However, GAN still has some critical shortcomings, for example, instability, and generating samples in aimless directions. To improve the performance of GAN, many competitive follow-up works have been developed. Jabbar *et al*. [23] has provided a detailed review of GAN-based methods and variants. For instance, Arjovsky, et al. [24] proposed Wasserstein GAN (WGAN) to improve the stability with applying Wasserstein distance instead of the Jensen–Shannon divergence (JS divergence). Mao *et al.* [25] also proposed least square GAN (LSGAN) to make the model more stable by applying least square entropy loss instead of cross-entropy loss. Besides, Radford *et al.* [26] proposed deep convolutional GAN (DCGAN) for learning unsupervised representation. Furthermore, in order to standardize the direction of generating samples, adding extra variables in the generator becomes a popular way. Among such type of GAN-based approaches, conditional GAN (CGAN) [27] is one of the most popular variants. It adds response variables, **y**, in the framework so that the directions to generate artificial samples in CGAN are restricted by the values of response variable. Based on CGAN, more advanced GAN-based models are developed. Brock *et al.* [28] also proposed Big Generative Adversarial Networks (BigGAN) for large scale training based on CGAN. It developed a conditional generator by applying the latent representations in multiple hidden layers of the generator. In addition, Karras *et al.* [29] demonstrated a style-based generator architecture for GAN. It applied the styles of images in the generator, similar as the labels. However, the approaches like CGAN for image augmentation usually consider the response variables as support rather than the expressions, which is not appropriate to be applied directly for this study.

Apart from the variants proposed for image augmentation, there are also some GAN-based methods developed in regression. For example, Rezagholiradeh *et al.* [30] proposed Reg-Gan to solve the semi-supervised learning problem. Its discriminator involves two tasks, predicting the labels and distinguishing the samples. In this way, it could not only update the model but also predict continuous labels rather than discretized labels. However, the response variables in Reg-Gan are usually relatively low dimension so that such approaches could not learn the relatively low dimension to high dimension mapping for this study. Furthermore, one of the most popular applications for GAN-based regression is image captioning [31]. For instance, Dai *et al.* [32] proposed a framework based on CGAN so that the generator can generate captions conditioned on images. Reinforcement learning is also applied to allow the generator to receive early feedback. Besides, Shetty *et al.* [33] also proposed a framework with applying two distances in the discriminator. One is for measuring the distance between image and captions while another is for measuring the distance between captions and captions. In addition, both GAN-based image captioning approaches mentioned above applied mismatching to guide the iteration direction. However, the applied mismatching is not sufficient because the pairs of wrong images and right captions are not involved.

To synthesize expressions in a GAN-based framework, Reed *et al* [9] proposed GAN-CLS, which is able to generate appropriate images based on given captions. Compared with image captioning approaches, this approach fully considers three pairs in the training process: (1) real images and right captions, (2) real images and wrong captions, as well as (3) fake images and right captions. However, the artificial data distributions are not able to be the same as actual data distributions since GAN-CLS is the biased estimation. Besides, its generation may still be instable.

Based on the above-mentioned problem statement and related literature review, the challenges of this paper could be summarized into three aspects: (1) the mapping from the relatively low dimension to relatively high dimension containing both deterministic and stochastic patterns, is hard to be learnt; (2) The existing GAN-based architecture which could be able to learn the mapping, may result in significant learning bias due to the mismatching errors; and (3) the training and generation process of the GAN-based architecture are still instable.

To address the challenges mentioned above, this paper developed a selective ensemble characteristic-to-expression synthesis (SE-CTES) approach. Since the characteristic-to-expression mapping contains both deterministic and stochastic patterns, a GAN-based modeling framework with deep neural networks is incorporated to learn the mapping following the architecture of GAN-CLS. In addition, the weights of mismatching errors are proposed to be different in model training. In this way, the trained data distribution is the domination in the estimation to the actual data distribution and the learning bias could be reduced. Besides, in order to improve the model robustness and further reduce the prediction bias, a novel selective ensemble framework is proposed.

## 3   Proposed Methodology

The overall framework of the proposed research methodology is summarized consisting of four aspects: (1) the proposed characteristic-to-expression synthesis (CTES) approach based on GAN is developed in Sec. 3.1; (2) the properties of proposed model are discussed in Sec. 3.2; (3) the proposed selective ensemble learning framework is presented in Sec. 3.3; and (4) model validation is described in Sec. 3.4.

### 3.1   Proposed characteristic-to-expression synthesis approach (CTES) based on GAN

Motivated by the GAN-CLS [9] reviewed in Sec. 2.2, the overall architecture of the proposed CTES is illustrated in Figure 2. It contains a generator $G: \mathbb{R}^z \times \mathbb{R}^m \rightarrow \mathbb{R}^n$ to generate **y** given **x** and a discriminator $D: \mathbb{R}^n \times \mathbb{R}^m \rightarrow \{0, 1\}$ to distinguish different pairs of **x** and **y**.

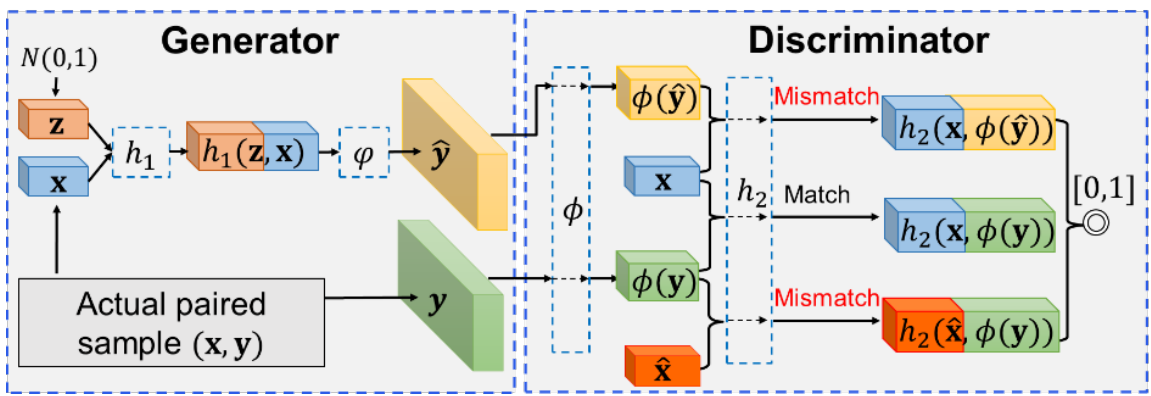

*Figure 2: A demonstration of CTES approach*

In $G$, the characteristic **x** is first mixed with a noise **z** randomly generated from a normal distribution $N(0,1)$. Subsequently, a fully connected neural network $h_1$ is applied to generate the feature vector of the expressions, $h_1(\mathbf{z}, \mathbf{x})$. Afterwards, a decoding network $\varphi$ is applied to synthesize $\hat{\mathbf{y}}$ by $h_1(\mathbf{z}, \mathbf{x})$, i.e., $\hat{\mathbf{y}} = \varphi(h_1(\mathbf{z}, \mathbf{x}))$. Hence, the generator $G$ for any input $\mathbf{z}, \mathbf{x}$ can be represented as shown in Eq. (4).

$$G(\mathbf{z}, \mathbf{x}) = \varphi(h_1(\mathbf{z}, \mathbf{x})) \qquad (4)$$

In $D$, another encoding network $\phi$ is applied to encode $\hat{\mathbf{y}}$ and $\mathbf{y}$ into feature vectors $\phi(\hat{\mathbf{y}})$ and $\phi(\mathbf{y})$, respectively, which is equivalent to $\varphi^{-1}$. The actual characteristic, $\mathbf{x}$, is sent to the discriminator $D$ (Figure 2). In addition, the unmatched actual characteristic from the data, $\hat{\mathbf{x}}$, which is randomly selected and does not match the actual expression $\mathbf{y}$, is considered as the fake characteristic and also sent to $D$. Afterwards, another fully connected neural network, $h_2$, is applied to discriminate the three pairs, $(\mathbf{x}, \mathbf{y})$, $(\mathbf{x}, \hat{\mathbf{y}})$, and $(\hat{\mathbf{x}}, \mathbf{y})$. The output of $h_2$ is a scalar in $[0,1]$. Therefore, the discriminator $D$ for any input $(\mathbf{x}, \mathbf{y})$ can be represented as Eq. (5).

$$D(\mathbf{x}, \mathbf{y}) = h_2(\mathbf{x}, \phi(\mathbf{y})) \qquad (5)$$

For convenience, the outputs of discriminator for $(\mathbf{x}, \mathbf{y})$, $(\mathbf{x}, \hat{\mathbf{y}})$, $(\hat{\mathbf{x}}, \mathbf{y})$ are represented by $D(\mathbf{x}, \mathbf{y})$, $D(\mathbf{x}, \hat{\mathbf{y}})$, $D(\hat{\mathbf{x}}, \mathbf{y})$, respectively.

It is important to note that the mismatching of $(\hat{\mathbf{x}}, \mathbf{y})$ and $(\mathbf{x}, \hat{\mathbf{y}})$ are completely different. $(\hat{\mathbf{x}}, \mathbf{y})$ is always mismatching while $(\mathbf{x}, \hat{\mathbf{y}})$ may switch from mismatching to matching. At the beginning of training, $(\mathbf{x}, \hat{\mathbf{y}})$ is also mismatching because $\hat{\mathbf{y}}$ is not accurate. Through the gradual training for the generator, $\hat{\mathbf{y}}$ is gradually approaching to $\mathbf{y}$ so that $(\mathbf{x}, \hat{\mathbf{y}})$ may also be considered as matching. In the GAN-CLS which inspires the proposed method, these two mismatching pairs are also applied but have the same weight [9]. However, given that these two mismatching pairs play different roles, the weights should be assigned accordingly. Therefore, $\beta$ is proposed to determine the weights of two mismatching errors as shown in the mini-max game to train the CTES model formulated as Eq.(6),

$$\min_{G} \max_{D} V(D, G) = \mathbb{E}_{(\mathbf{x},\mathbf{y}) \sim P_{\text{data}}}[\log(D(\mathbf{x}, \mathbf{y}))] + \beta \mathbb{E}_{(\mathbf{x},\hat{\mathbf{y}}) \sim P_{G(\mathbf{z},\mathbf{x})}}[\log(1 - D(\mathbf{x}, \hat{\mathbf{y}}))]$$
$$+ (1 - \beta) \mathbb{E}_{(\hat{\mathbf{x}},\mathbf{y}) \sim P'_{\text{data}}}[\log(1 - D(\hat{\mathbf{x}}, \mathbf{y}))] \qquad (6)$$

where $\beta \in \{0,1\}$ is the weight of the mismatch $(\mathbf{x}, \hat{\mathbf{y}})$ and $1-\beta$ is the weight of mismatch $(\hat{\mathbf{x}}, \mathbf{y})$. $(\mathbf{x}, \mathbf{y})$ follows the joint distribution of data, namely, $P_{\text{data}}$. $(\hat{\mathbf{x}}, \mathbf{y})$ follows the joint distribution that has the incorrect $\mathbf{x}$, i.e., $P'_{\text{data}}$, and $(\mathbf{x}, \hat{\mathbf{y}})$ follows the joint distribution based on $G$, $P_{G(\mathbf{z},\mathbf{x})}$. All the related distributions, $P_{\text{data}}, P_{G(\mathbf{z},\mathbf{x})}$, and $P'_{\text{data}}$, are unknown. However, it will not influence the iteration process since the samples, i.e., $(\mathbf{x}, \mathbf{y}), (\mathbf{x}, \hat{\mathbf{y}}), (\hat{\mathbf{x}}, \mathbf{y})$, could still be obtained from the distributions. As $P_{G(\mathbf{z},\mathbf{x})}$ is close to $P_{\text{data}}$, the synthesized expressions will be similar as actual expressions. Then the objective to learn the characteristic-to-expression synthesis could be achieved. Hence, $P_{G(\mathbf{z},\mathbf{x})}, P_{\text{data}}$ and $P'_{\text{data}}$ are not required to be pre-defined in the proposed method.

Among these three pairs, only $(\mathbf{x}, \hat{\mathbf{y}})$ is related to $G$. Then in Eq. (6), only $\beta \mathbb{E}_{(\mathbf{x},\hat{\mathbf{y}}) \sim P_{G(\mathbf{z},\mathbf{x})}}[\log(1 - D(\mathbf{x}, \hat{\mathbf{y}}))]$ is related to $G$ so that the loss for $G$, i.e., $L_G$, is only based on $(\mathbf{x}, \hat{\mathbf{y}})$. Besides, the loss for $D$, i.e., $L_D$, needs to consider all three pairs in Eq. (6). Hence, $L_D$ and $L_G$ can be demonstrated by Eq. (7). With the losses updated, $h_1, h_2, \varphi, \phi, D$ can also be updated simultaneously so that the model can be sufficiently trained after the losses converge.

$$L_D = \log(D(\mathbf{x}, \mathbf{y})) + \beta \log(1 - D(\mathbf{x}, \hat{\mathbf{y}})) + (1-\beta) \log(1 - D(\hat{\mathbf{x}}, \mathbf{y}))$$
$$L_G = \log(D(\mathbf{x}, \hat{\mathbf{y}}))$$
(7)

In practice, the synthesized expressions given the same $\mathbf{x}$ may lack diversity and concentrate on one specific area. Hence, in order to increase the diversity of synthesized expressions, one solution is to apply the extensions of $\mathbf{x}$ rather than only $\mathbf{x}$ for synthesis. For instance, when performing the synthesis, some noise could be added in $\mathbf{x}$ so that the expressions are synthesized under different $\mathbf{x}$. To assist the convergence of the proposed CTES model, the number of iterations could set as a sufficiently large number. Through multiple experimental trials, the number of iterations, $t$, could be set as 1,000.

Overall, the algorithm of training CTES is shown in Figure 3. In each iteration, two different groups of characteristics, $\mathbf{X}$ and $\hat{\mathbf{X}}$, are batched. It is important to note that, the fake characteristics, $\hat{\mathbf{X}}$, are randomly selected from the unmatched characteristics. The expressions **Y** are also taken according to **X**. Then $G$ transforms the input to fake expressions $\hat{\mathbf{y}}$. Afterwards, $D$ calculates the outputs for the three different pairs.

---

**Algorithm 1:** CTES algorithm

---
**Input:** Characteristics $\mathbf{X}_{\text{data}}$ with corresponding expressions $\mathbf{Y}_{\text{data}}$, number of iterations $t$, batch size $s$, $i = 0$
**For** $i = 1$ **to** $t$ **do:**
  **Step 1:** Batch characteristics $\mathbf{X} = \{\mathbf{x}^{(1)}, \mathbf{x}^{(2)}, \ldots, \mathbf{x}^{(s)}\}$ with corresponding expressions $\mathbf{Y} = \{\mathbf{y}^{(1)}, \mathbf{y}^{(2)}, \ldots, \mathbf{y}^{(s)}\}$
  **Step 2:** Randomly pick fake characteristics $\hat{\mathbf{X}} = \{\hat{\mathbf{x}}^{(1)}, \hat{\mathbf{x}}^{(2)}, \ldots, \hat{\mathbf{x}}^{(s)}\}$ from $\mathbf{X}_{\text{data}}$
  **For** $j = 1$ **to** $s$ **do:**
    In generator $G$:
      **Step 3:** Generate random noise input $\mathbf{z} \sim N(0,1)$
      **Step 4:** Generate output of $h_1$: $h_1(\mathbf{z}, \mathbf{x}^{(j)})$
      **Step 5:** Generate fake expressions $\hat{\mathbf{y}} = \varphi(h_1(\mathbf{z}, \mathbf{x}^{(j)}))$ based on the decoding CNN $\varphi$
    In discriminator $D$:
      **Step 6:** Apply $\phi$ to encoding $\mathbf{y}$ and $\hat{\mathbf{y}}$ to feature vectors $\phi(\mathbf{y})$ and $\phi(\hat{\mathbf{y}})$, respectively
      **Step 7:** Apply $h_2$ to obtain $h_2(\mathbf{x}^{(j)}, \phi(\mathbf{y}))$ as $D(\mathbf{x}, \mathbf{y})$
      **Step 8:** Apply $h_2$ to obtain $h_2(\hat{\mathbf{x}}^{(j)}, \phi(\mathbf{y}))$ as $D(\hat{\mathbf{x}}, \mathbf{y})$
      **Step 9:** Apply $h_2$ to obtain $h_2(\mathbf{x}^{(j)}, \phi(\hat{\mathbf{y}}))$ as $D(\mathbf{x}, \hat{\mathbf{y}})$
      **Step 10**: Update $L_G, L_D$
**If** $L_G, L_D$ converge**:**
  **Break and output** $G, D$;

---

*Figure 3: A demonstration of proposed CTES algorithm*

Though expressions **y** is assumed as high dimension, it is worth noting that the method could also be applied under low dimensional **y**. The property of the model also holds the same no matter what the dimension of **y** is, which is demonstrated in Sec. 3.2.

## 3.2 Model property

This section demonstrates the model properties of the proposed CTES model. First of all, the most important property is the convergence, which is demonstrated in proposition 1. When $\beta P_{G(\mathbf{z},\mathbf{x})} + (1-\beta)P'_{\text{data}} = P_{\text{data}}$ satisfies, the global minimum value of $\max_D V(D, G)$ can be achieved [8].

***Proposition 1.*** If $(\mathbf{x}, \mathbf{y})$ is from distribution $P_{\text{data}}$, $(\hat{\mathbf{x}}, \mathbf{y})$ is from distribution $P'_{\text{data}}$, and $(\mathbf{x}, \hat{\mathbf{y}})$ is from distribution $P_{G(\mathbf{z},\mathbf{x})}$, then the training will converge when $\beta P_{G(\mathbf{z},\mathbf{x})} + (1-\beta)P'_{\text{data}} = P_{\text{data}}$.

*Proof.* Essentially, if $G$ is settled down to maximize $D$, $V(G, D)$ can be transformed as following:

$$V(D,G) = \int_{(\mathbf{x},\mathbf{y})} P_{\text{data}}((\mathbf{x},\mathbf{y}))\left[\log(D(\mathbf{x},\mathbf{y}))\right]d(\mathbf{x},\mathbf{y}) + \int_{(\mathbf{x},\hat{\mathbf{y}})} \beta P_{G(\mathbf{z},\mathbf{x})}((\mathbf{x},\hat{\mathbf{y}}))\left[\log(1-D(\mathbf{x},\hat{\mathbf{y}}))\right]d(\mathbf{x},\hat{\mathbf{y}})$$

$$+ \int_{(\hat{\mathbf{x}},\mathbf{y})} (1-\beta)P'_{\text{data}}((\hat{\mathbf{x}},\mathbf{y}))\left[\log(1-D(\hat{\mathbf{x}},\mathbf{y}))\right]d(\hat{\mathbf{x}},\mathbf{y})$$

$$= \int_{(\mathbf{x},\mathbf{y})} P_{\text{data}}\left[\log(D(\mathbf{x},\mathbf{y}))\right] + (\beta P_{G(\mathbf{z},\mathbf{x})} + (1-\beta)P'_{\text{data}})\left[\log(1-D(\mathbf{x},\mathbf{y}))\right]d(\mathbf{x},\mathbf{y})$$

Hence, when $G$ is fixed, the optimal discriminator $D$ is shown as

$$D_G^*(\mathbf{x}) = \frac{P_{\text{data}}}{P_{\text{data}} + (\beta P_{G(\mathbf{z},\mathbf{x})} + (1-\beta)P'_{\text{data}})} \tag{8}$$

Therefore, the minimax game when $G$ is fixed is

$$\max_D V(D,G) = \mathbb{E}_{(\mathbf{x},\mathbf{y}) \sim P_{\text{data}}}\left[\log \frac{P_{\text{data}}}{P_{\text{data}} + (\beta P_{G(\mathbf{z},\mathbf{x})} + (1-\beta)P'_{\text{data}})}\right] + (\beta \mathbb{E}_{(\mathbf{x},\mathbf{y}) \sim P_{G(\mathbf{z},\mathbf{x})}}$$
$$+ (1-\beta)\mathbb{E}_{(\mathbf{x},\mathbf{y}) \sim P'_{\text{data}}})\left[\log \frac{\beta P_{G(\mathbf{z},\mathbf{x})} + (1-\beta)P'_{\text{data}}}{P_{\text{data}} + (\beta P_{G(\mathbf{z},\mathbf{x})} + (1-\beta)P'_{\text{data}})}\right] \tag{9}$$

Hence, when Eq. (10) satisfies, $\max_D V(D,G)$ could achieve its global minimum value $-\log 4$ [6].

$$\beta P_{G(\mathbf{z},\mathbf{x})} + (1-\beta)P'_{\text{data}} = P_{\text{data}} \tag{10}$$

Q.E.D

If $\beta = 1$, another mismatching from $P'_{\text{data}}$ disappears so that the global minimum value of $\max_D V(D,G)$ can be achieved by $P_{G(\mathbf{z},\mathbf{x})} = P_{\text{data}}$. That is, $P_{G(\mathbf{z},\mathbf{x})}$ is an unbiased estimation for $P_{\text{data}}$. However, the unbiased estimation still cannot avoid the model collapse issue so that it may not be the best in practice. Thus, when $\beta < 1$, though $P_{G(\mathbf{z},\mathbf{x})}$ will become the biased estimation for $P_{\text{data}}$, $P'_{\text{data}}$ could be applied for guiding the iteration direction. To discuss the effects of $\beta$ on the learning bias in the training process, according to the Proposition 1, the Corollary 1 is demonstrated below.

***Corollary 1.*** To make $P_{G(\mathbf{z},\mathbf{x})}$ robustly approximate to $P_{\text{data}}$, $\beta$ should be larger than 0.5.

***Proof***: According to the Eq. (10), $\beta P_{G(z,x)} + (1-\beta)P'_{data}$ may be approximate to $P_{data}$ when the iteration converges. Thus, given $\beta$ and $P_{data}$, the increments of $d(P_{data}, P'_{data})$ will also result in the increments of $d(P_{data}, P_{G(z,x)})$ simultaneously. However, the goal is to make $P_{G(z,x)}$ robustly approximate to $P_{data}$, i.e., make $P_{G(z,x)}$ robustly approximate to the mixture distribution. Hence, the mixture distribution, $\beta P_{G(z,x)} + (1-\beta)P'_{data}$, should be dominated by $P_{G(z,x)}$, i.e., $\beta > (1-\beta)$, which indicates $\beta > 0.5$.

Q.E.D.

In order to better illustrate Corollary 1, Figure 4 demonstrates it in a graphical way. $P_{G(z,x)}$ (blue, solid line) could be seen as a mixture distribution of both $P_{data}$ (brown, dashed line) and $P'_{data}$ (black, solid line). The ideal situation is shown in Figure 4(a) that $P_{G(z,x)}$ is similar to $P_{data}$ and $P_{G(z,x)}$ is considered as a well-trained distribution. However, as $d(P_{data}, P'_{data})$ increases, the situation in Figure 4(b) may also happen that $P_{G(z,x)}$ is far away from $P_{data}$. Under such circumstances, though the iterations may also converge, $P_{G(z,x)}$ is not close to $P_{data}$. Hence, given that $P'_{data}$ is not stable, $P_{G(z,x)}$ should be the domination of the mixture distribution $\beta P_{G(z,x)} + (1-\beta)P'_{data}$, i.e., $\beta > 0.5$.

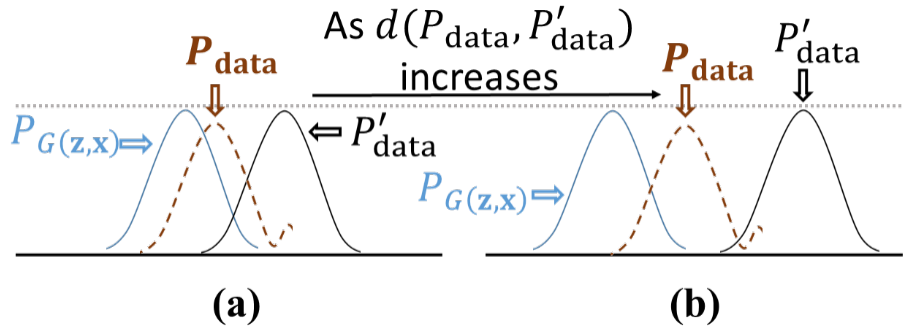

***Figure 4:*** *A demonstration of different kinds of samples. Essentially, $P_{data}$ (brown, dashed line) is settled down while $P'_{data}$ (black, solid line) may be not fixed. Therefore, $P_{G(z,x)}$ (blue, solid line) may differ a lot. (a) is an ideal situation so that most of the generated samples are also located in the area of $P_{data}$, but (b) is the situation that the estimation is far away from the actual distribution, $P_{data}$. Therefore, the inspection and selection for generated distribution is further needed.*

In practice, the exact value of $\beta$ could be determined by cross validation. However, the instable training process for CTES is essential to be solved. A natural way to solve this problem is to train multiple models and select the models with closer $P_{G(z,x)}$ to $P_{data}$, i.e., the models that can generate accurate expressions. Following this direction, a selective ensemble framework as described in Sec. 3.3 is proposed to improve the robustness of CTES.

### 3.3 Selective ensemble learning architecture

According to the Eq. (10) and Corollary 1, making $P_{G(z,x)}$ dominate the mixture distribution, $\beta P_{G(z,x)} + (1-\beta)P'_{data}$, could enable a relatively robust approximation. However, in practice, considering a single round of training, as shown in Figure 5, due to the randomness of $P'_{data}$, it may still be hard to completely guarantee a satisfactory approximation of $P_{G(z,x)}$ to $P_{data}$ if $d(P_{data}, P'_{data})$ is very large. To further address this issue, a natural idea is to perform multiple rounds of training (i.e., obtain multiple $P_{G(z,x)}$) and selectively ensemble the good $P_{G(z,x)}$ (i.e., with small $d(P_{data}, P_{G(z,x)})$). Therefore, the selective ensemble learning framework is proposed in this work to reduce the prediction bias and improve the synthesis stability. Its potential and feasibility can be justified by the Corollary 2.

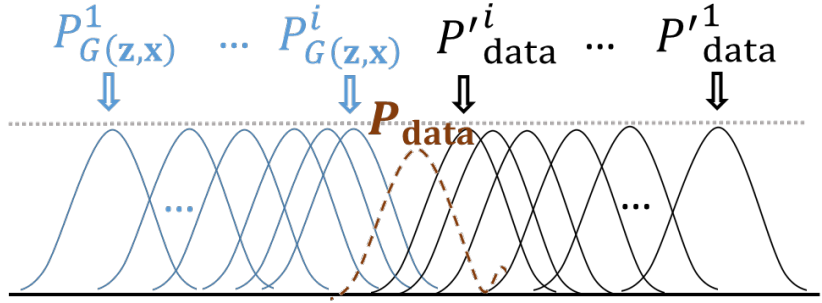

**Figure 5:** *Demonstrations of the relative distance among $P_{G(z,x)}$ (blue, solid line), $P_{data}$ and $P'_{data}$ (black, solid line) of different CTES models*

***Corollary 2.*** The mixture of the selected $P_{G(z,x)}$ with small $d(P_{data}, P_{G(z,x)})$, is denoted as $MP_{G(z,x)}$. When the expectations of selected $P_{G(z,x)}$, $\mathbb{E}(P_{G(z,x)})$, are approaching to the expectation of $P_{data}$, $\mathbb{E}(P_{data})$, the expectation of $MP_{G(z,x)}$, $\mathbb{E}(MP_{G(z,x)})$, gradually converges to $\mathbb{E}(P_{data})$.

***Proof***: For any $\varepsilon > 0$, given that $\mathbb{E}(P_{G(z,x)})$ is approaching to $\mathbb{E}(P_{data})$, many $G(z, x)$ could be selected to satisfy the condition that $\left|\mathbb{E}(P_{G(z,x)}) - \mathbb{E}(P_{data})\right| < \varepsilon$. Denote the selected $P_{G(z,x)}$ as $P^{\{1\}}_{G(z,x)}, P^{\{2\}}_{G(z,x)}, \ldots,$

$P_{G(\mathbf{z},\mathbf{x})}^{\{h\}}$. $MP_{G(\mathbf{z},\mathbf{x})}$ is the weighted average of selected $P_{G(\mathbf{z},\mathbf{x})}$, i.e., $MP_{G(\mathbf{z},\mathbf{x})} = \sum_{i=1}^{h} \frac{1}{h} P_{G(\mathbf{z},\mathbf{x})}^{\{i\}}$. Then Eq. (11) is obtained. Following the definition of limit, $\mathbb{E}(MP_{G(\mathbf{z},\mathbf{x})})$ converges to $\mathbb{E}(P_{\text{data}})$.

$$\left| \mathbb{E}(MP_{G(\mathbf{z},\mathbf{x})}) - \mathbb{E}(P_{\text{data}}) \right| = \left| \mathbb{E}\left( \sum_{i=1}^{h} \frac{1}{h} P_{G(\mathbf{z},\mathbf{x})}^{\{i\}} \right) - \mathbb{E}(P_{\text{data}}) \right| = \left| \sum_{i=1}^{h} \frac{1}{h} \mathbb{E}\left( P_{G(\mathbf{z},\mathbf{x})}^{\{i\}} \right) - \mathbb{E}(P_{\text{data}}) \right|$$
$$< \sum_{i=1}^{h} \frac{1}{h} \left| \mathbb{E}\left( P_{G(\mathbf{z},\mathbf{x})}^{\{i\}} \right) - \mathbb{E}(P_{\text{data}}) \right| < \sum_{i=1}^{h} \frac{1}{h} \varepsilon = \varepsilon \tag{11}$$

Q.E.D.

Figure 6 demonstrates corollary 2 in a graphical way. Some of the $P_{G(\mathbf{z},\mathbf{x})}$ (blue, solid line) may be close to $P_{\text{data}}$ (brown, dashed line), while others may be far away. Then $P_{G(\mathbf{z},\mathbf{x})}$ with small $d(P_{\text{data}}, P_{G(\mathbf{z},\mathbf{x})})$ in the red rectangle could be selected. Afterwards, the selected $P_{G(\mathbf{z},\mathbf{x})}$ could be ensembled as a mixture distribution, $MP_{G(\mathbf{z},\mathbf{x})}$ (red, solid line). Given that the selected $P_{G(\mathbf{z},\mathbf{x})}$ may locate on both sides of $P_{\text{data}}$, The expectation of $MP_{G(\mathbf{z},\mathbf{x})}$ is closer to the expectation of $P_{\text{data}}$, i.e., $\mathbb{E}(MP_{G(\mathbf{z},\mathbf{x})})$ could converge to $\mathbb{E}(P_{\text{data}})$. Hence, the selective ensemble learning architecture could be applied to select an optimal set of $P_{G(\mathbf{z},\mathbf{x})}$, i.e., an optimal set of the trained CTES models.

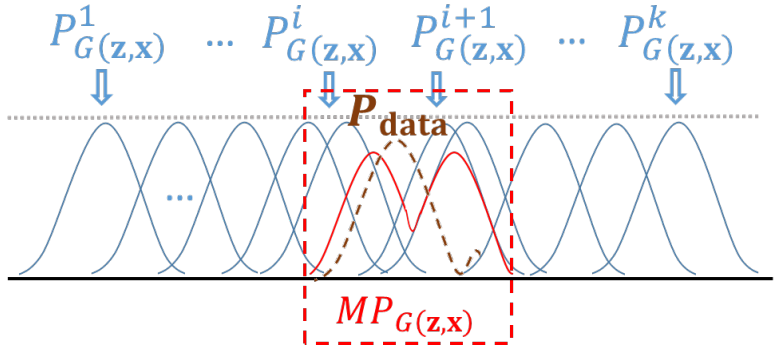

**Figure 6:** *A demonstration of different $P_{G(\mathbf{z},\mathbf{x})}$ (blue, solid line) given $P_{data}$ (brown, dashed line). Then a mixture of selected $P_{G(\mathbf{z},\mathbf{x})}$, $MP_{G(\mathbf{z},\mathbf{x})}$, (red, solid line) could be selected and ensembled as a less biased estimation than each individual $P_{G(\mathbf{z},\mathbf{x})}$.*

Notably, a practical to implement the selective ensemble learning is that the CTES models are hard to be compared directly for selection. To address this issue, a natural idea is to validate the effectiveness of the trained CTES models to select the qualified CTES models. To achieve that, the synthesized expressions and their corresponding characteristics to generate expressions are inversely correlated based on a supervised machine learning model, i.e., the inverse validation model. If the synthesized expressions are effective, the inverse validation framework should be able to output the correct corresponding characteristics which synthesized the expressions. In this way, the evaluation metrics for the inverse validation framework could demonstrate the quality of the synthesized expressions from each CTES model. Then the CTES models with better performance in the inverse validation framework are selected. The detailed definition of the proposed inverse validation model is shown in definition 1.

***Definition 1.*** (**Inverse validation model**): The inverse validation model, $M$, applies the generated expressions from CTES to validate the effectiveness of CTES. Assume the CTES model is trained $k$ times using $(\mathbf{x}, \mathbf{y})$: $\{(\mathbf{x}^{(1)}, \mathbf{y}^{(1)}), \ldots, (\mathbf{x}^{(N)}, \mathbf{y}^{(N)})\}$. Afterwards, fake expressions $\hat{\mathbf{y}}$: $\{\hat{\mathbf{y}}^{(1)}, \ldots, \hat{\mathbf{y}}^{(k)}\}$ are synthesized from the $k$ fitted CTES models. Then there requires $k$ inverse validation models, $M_1, \ldots, M_k$. In $M_i$, i.e., the $i$-th inverse validation model, $\{\hat{\mathbf{y}}^{(1)}, \ldots, \hat{\mathbf{y}}^{(i-1)}, \hat{\mathbf{y}}^{(i+1)} \ldots, \hat{\mathbf{y}}^{(k)}\}$ are combined as positive expressions and $\{\mathbf{y}^{(1)}, \mathbf{y}^{(2)}, \ldots, \mathbf{y}^{(n)}\}$ are combined as negative expressions in the training set. Meanwhile, the $i$-th expression in $\hat{\mathbf{y}}$, i.e., $\hat{\mathbf{y}}^{(i)}$, is applied as the testing set for $M_i$. Finally, $M_i$ outputs its accuracy $a_i$.

As described in Figure 6, not all $P_{G(\mathbf{z},\mathbf{x})}$ trained by CTES models are close to $P_{\text{data}}$. Thus, the optimal set of CTES models should be selected by the synthesized expressions according to the inverse validation models. If $\hat{\mathbf{y}}^{(i)}$ is considered as an accurate expression, $M_i$ should classify $\hat{\mathbf{y}}^{(i)}$ as positive expressions rather than negative expressions. Hence, the accuracy of $M_i$, i.e., $a_i$, could be considered as the quality of $\hat{\mathbf{y}}^{(i)}$. In this way, $M_i$ could evaluate the effectiveness of the $i$-th CTES model.

Based on the inverse validation models, a demonstration of SE-CTES framework is shown in Figure 7 and the algorithm of SE-CTES is described in Figure 8. Initially, CTES models are trained $k$ times as $\hat{f}_1, \ldots, \hat{f}_k$. Then based on $\hat{\mathbf{y}}$ and $\mathbf{y}$, $k$ inverse validation models, $M_1, \ldots, M_k$, could be trained and tested. The accuracy of these $k$ inverse validation model, $a_1, a_2, ..., a_k$, could be obtained. Afterwards, $h$ CTES models with the highest accuracy, $\{\hat{f}_1', \hat{f}_2', ..., \hat{f}_h'\}$, are selected and combined together as the desired ensemble model.

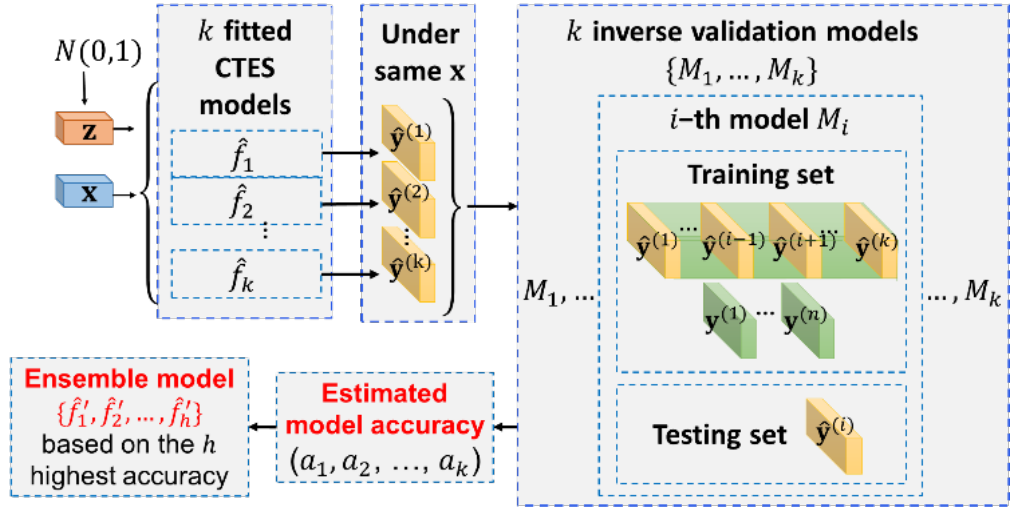

*Figure 7: The procedure of the proposed SE-CTES framework*

| Algorithm 2: The proposed SE-CTES algorithm |
| --- |
| **Input:** Characteristics $\mathbf{x}$, existing expressions $\mathbf{y} = \{\mathbf{y}^{(1)}, \mathbf{y}^{(2)}, ..., \mathbf{y}^{(n)}\}$, parameter $k$, $h$ <br> Randomly generate noise $\mathbf{z}$ <br> **For** $i$ **in** 1 **to** $k$ **do:** <br>     **Step 1:** Train the CTES model $\hat{f}_i$ based on **algorithm 1** <br>     **Step 2:** Generate samples $\hat{\mathbf{y}}^{(i)}$ with inputting $\mathbf{x}$ and $\mathbf{z}$ to the trained model $\hat{f}_i$ <br> **For** $i$ **in** 1 **to** $k$ **do:** <br>     **Step 3:** Combine $\{\hat{\mathbf{y}}^{(1)}, ..., \hat{\mathbf{y}}^{(i-1)}, \hat{\mathbf{y}}^{(i+1)} ..., \hat{\mathbf{y}}^{(k)}\}$ together as category 0 in the training set <br>     **Step 4:** Send $\mathbf{y}$ into the training set as category 1 <br>     **Step 5:** Train the inverse validation model $M_i$ based on the training set <br>     **Step 6:** Pick $\hat{\mathbf{y}}_i$ as the testing set for $M_i$ and output the accuracy $a_i$ of testing set <br> **Step 7:** Rank $\{a_1, a_2, a_3, ..., a_n\}$ and pick the first $h$ corresponding CTES models $\{\hat{f}_1', \hat{f}_2', ..., \hat{f}_h'\}$ as the ensemble model |

*Figure 8: A demonstration of the proposed SE-CTES algorithm*

## 3.4 Model validation

According to the previous sections, the proposed SE-CTES could build an ensemble model for synthesis based on the selective ensemble learning framework. It is also important to validate the effectiveness of the ensemble model, i.e., the effectiveness of the proposed SE-CTES. A natural idea is to check whether the synthesized expressions are similar as actual expressions. If the synthesized expressions are similar as actual expression, the proposed SE-CTES is effective. Thus, one common way for validation is to calculate the differences between synthesized expressions and actual expressions. In practice, such validation could be applied when the expressions are low dimension. However, when expressions are high dimension or more complex, the differences between synthesized expressions and actual expressions may not be convincing. Hence, as shown in Figure 9, this section demonstrates a validation framework for high dimension expressions. It is important to note that, the classification is applied to validate the quality of synthesized expressions in the proposed validation framework. If synthesized expressions are accurate, then the classification performance (e.g., classification accuracy) should be better. In this way, the classification metric could demonstrate whether the proposed method could synthesize the expressions accurately or not.

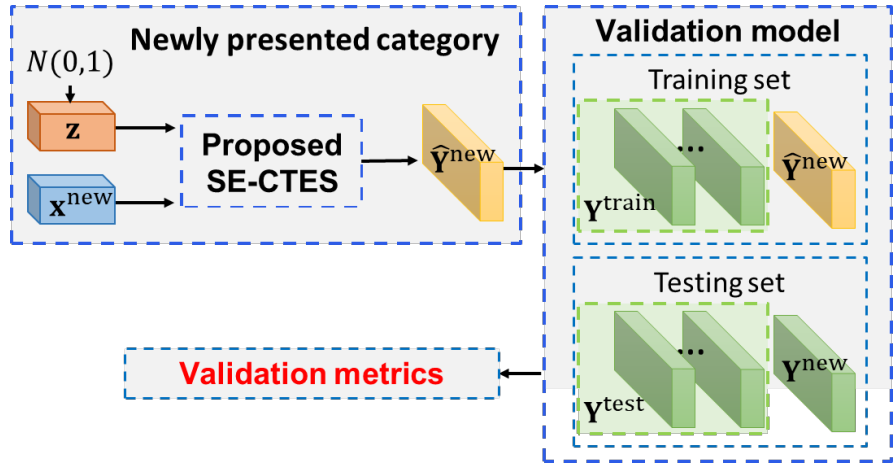

*Figure 9: A demonstration of applied validation framework*

Assume that $(\mathbf{X}^{(1)}, \mathbf{Y}^{(1)}), (\mathbf{X}^{(2)}, \mathbf{Y}^{(2)}), \ldots, (\mathbf{X}^{(c)}, \mathbf{Y}^{(c)})$ have different $c$ groups labeled by characteristics. Denote $(\mathbf{X}^{\text{new}}, \mathbf{Y}^{\text{new}})$ as a newly presented pair which belongs to a new group. The model is to synthesize $\hat{\mathbf{Y}}^{\text{new}}$ based on $\mathbf{X}^{\text{new}}$. Then the comparison between $\mathbf{Y}^{\text{new}}$ and $\hat{\mathbf{Y}}^{\text{new}}$ is considered as the validation criteria. Initially, 50% expressions of $\mathbf{Y}^{(1)}, \mathbf{Y}^{(2)}, \ldots, \mathbf{Y}^{(c)}$ are extracted as $\mathbf{Y}^{\text{train}}$ according to categories, and the other 50% expressions are left as $\mathbf{Y}^{\text{test}}$. $\mathbf{Y}^{\text{train}}$ is incorporated with $\hat{\mathbf{Y}}^{\text{new}}$ as the training set in the validation model, and $\mathbf{Y}^{\text{test}}$ is incorporated with $\mathbf{Y}^{\text{new}}$ as the testing set. If the pattern of $\hat{\mathbf{Y}}^{\text{new}}$ is consistent with $\mathbf{Y}^{\text{new}}$, $\mathbf{Y}^{\text{new}}$ should be classified into the group of $\hat{\mathbf{Y}}^{\text{new}}$, and $\mathbf{Y}^{\text{test}}$ should not be classified into the group of $\hat{\mathbf{Y}}^{\text{new}}$. Hence, the validation metric for such two situations could be applied to validate the quality of $\hat{\mathbf{Y}}^{\text{new}}$. If the validation metric (e.g., classification accuracy) is high, $\hat{\mathbf{Y}}^{\text{new}}$ is similar with $\mathbf{Y}^{\text{new}}$, meaning that the proposed SE-CTES is effective. In this way, the effectiveness of the proposed model could be validated.

## 4 Case Studies

In this section, two case studies are conducted to validate the proposed method based on the multivariate data. The first case conducted in Sec. 4.1 is a numerical simulation study where the simulated characteristics could be applied as labels for expressions in validation. Hence, the model validation framework in Sec. 3.4 could be applied to demonstrate the effectiveness of proposed method. The second case study performed in Sec. 4.2 is a real-world healthcare study for cardiovascular disease (CVD) risk prediction. To achieve more accurate prediction, there is a strong need of effective synthesis for the expressions (i.e., the physiological symptoms). For both studies, the key validation criterion is to validate whether the synthesized expressions are similar as actual expressions.

### 4.1 Simulation study

#### 4.1.1 Data generation and experimental setups

The effectiveness of proposed method is first validated by a simulation study. In the simulation data, five groups of characteristics are initially generated, namely, $\mathbf{X}^{(1)}, \mathbf{X}^{(2)}, \mathbf{X}^{(3)}, \mathbf{X}^{(4)}, \mathbf{X}^{(5)}$. Each characteristic $\mathbf{x}$ contains two variables, $(x_1, x_2)$, and $x_1, x_2$ in group $i$ satisfy the Eq.(12),

$$x_1, x_2 \sim N(\mu, \sigma^2),$$
$$\mu = 0.2i, \quad i = 1,2,3,4,5 \tag{12}$$

where $\sigma$ is a tuning parameter. Each group contains 200 samples.

Afterwards, as shown in Eq. (13), the expressions $\mathbf{y}$, which contains six variables, were then generated based on $\mathbf{x}$. Thus, in this case, $m = 2$ and $n = 6$. The first three variables in $\mathbf{y}$ are the first three orders Taylor polynomial of the Gaussian kernel function for $x_1$ and $x_2$ at the origin (0,0), while the other three variables are generated by squares of $x_1$ and $x_2$, as well as their multiplication. Besides, to introduce stochastic patterns in the expressions, noises are also added in the variable transformations. The expressions also belong to five groups, $\mathbf{Y}^{(1)}, \mathbf{Y}^{(2)}, \mathbf{Y}^{(3)}, \mathbf{Y}^{(4)}, \mathbf{Y}^{(5)}$.

$$\mathbf{Y}^{(i)} = \begin{bmatrix} \mathbf{y}_{i,1} \\ \vdots \\ \mathbf{y}_{i,200} \end{bmatrix}, \mathbf{y}_{i,j} = [y_1, y_2, y_3, y_4, y_5, y_6]^T, \mathbf{x}_{i,j} = [x_1, x_2]^T, \varepsilon \sim N(0, 0.005^2)$$

$$y_m = \frac{2^{m-1}}{(m-1)!}(x_1 + \varepsilon)^{m-1}(x_2 + \varepsilon)^{m-1} e^{-(x_1+\varepsilon)^2 - (x_2+\varepsilon)^2}, m = 1,2,3 \qquad (13)$$

$$y_4 = (x_1 + \varepsilon)^2, y_5 = (x_2 + \varepsilon)^2, y_6 = (x_1 + \varepsilon)(x_2 + \varepsilon)$$

$$i = 1,2,3,4,5, \qquad j = 1,2,\ldots,200$$

Based on the simulation data, $\mathbf{Y}^{(2)}, \mathbf{Y}^{(3)}, \mathbf{Y}^{(4)}$ are separately identified for model validation. The model validation follows the framework presented in Figure 9 of Sec. 3.4. When identifying group $i$, the proposed SE-CTES is first trained based on the other four groups to synthesize $\hat{\mathbf{Y}}^{(i)}$ based on $\mathbf{X}^{(i)}$. Afterwards, a classifier is trained by $\hat{\mathbf{Y}}^{(i)}$ and $\mathbf{Y}^{\text{train}}$. Then the classifier is applied to classify $\mathbf{Y}^{(i)}$ and $\mathbf{Y}^{\text{test}}$. Two accuracies, $A_1$ and $A_2$, are applied to evaluate the performance of proposed method based on two perspectives, which are explained in Figure 10. $A_1$ demonstrates the ratio of $\mathbf{Y}^{(i)}$ that is classified into group $i$ to entire $\mathbf{Y}^{(i)}$, while $A_2$ demonstrates the ratio of $\mathbf{Y}^{\text{test}}$ that is classified into group $1,\ldots,i-1,i+1,\ldots,5$ to entire $\mathbf{Y}^{\text{test}}$. In this way, if the proposed method is effective, the testing set should be classified correctly. Hence, the higher both $A_1$ and $A_2$ are, the better.

|  |  | Apply $\hat{\mathbf{Y}}^{(i)}$ in the training set | |
|---|---|---|---|
|  |  | Group $i$ | Group $1,\ldots, i-1, i+1,\ldots, 5$ |
| Apply $\mathbf{Y}^{(i)}$ in the testing set | Group $i$ | True positive (TP) | False positive (FP) |
|  | Group $1,\ldots,i-1, i+1,\ldots,5$ | False negative (FN) | True negative (TN) |

$$A_1 = \frac{TP}{TP + FP}$$

$$A_2 = \frac{TN}{TN + FN}$$

***Figure 10:*** *A demonstration of the calculation for both $A_1$ and $A_2$ when identifying group $i$*

In this study, the random forest classifier [34] is selected in both inverse validation and model validation. Both $D$ and $G$ use multilayer perceptron (MLP) and the depth of $D$ and $G$ is set to 3. The last layer in $G$ does not apply any activation function while the last layer in $D$ applies the sigmoid activation function. The other layers in $D$ and $G$ utilize the common ReLu function as the activation function. Besides, the batch size $s$ is set as 50.

The hyper-parameter $\beta$ is a tuning parameter. In this study, it is tuned manually among $\{0.5, 0.6, 0.7, 0.8, 0.9\}$. Each value from such set is utilized as $\beta$ to train the model, separately. Then the actual expressions are sent to the model validation framework as the testing set to output the validation metric. The $\beta$ with the best performance in terms of the evaluation metric is applied in the following studies. Hence, $\beta = 0.9$ is utilized. Besides, Bayesian optimization is also an alternative way to tune this parameter.

To avoid the data imbalanced issue, the number of synthesized expressions should be consistent with the number of actual expressions. Hence, $h$ is set as 2 to make the number of synthesized expressions equal to actual expressions. Then 100 synthesized expressions should be chosen from the selective ensemble learning framework. Based on the model property described in Sec. 3.3, more than half of the CTES models should have good estimations to $P_{\text{data}}$ (i.e., they are well trained). In order to make sure the selected CTES models are well trained, $k$ should be larger than $2h$. Due to the computational cost, $k$ is set as 5. As described in Sec. 3.1, the unmatched characteristics are randomly selected from the actual characteristics as fake characteristics.

To further evaluate the effectiveness of SE-CTES, benchmark approaches are also applied for comparisons. Since this problem can be formulated by regression, PLS regression [16] and GRNN [17] are selected as benchmark methods. The GAN-CLS [9] is also selected as one of the benchmark methods since the proposed SE-CTES is inspired by this approach. Besides, the CTES without selective ensemble is applied as a benchmark method as well to validate the effectiveness of the proposed selective ensemble learning framework. Furthermore, as a widely applied GAN method, "CGAN" is also selected. Notably, to fit the objective in this study, the "CGAN" applied here is not the same as CGAN in [27] because the input for the "CGAN" here is characteristics and expressions instead of the expressions and labels. All the applied GAN-based benchmark methods have the same setup with the proposed method.

4.1.2 Results and discussions

Five values of parameter $\sigma$, including {0.01, 0.03, 0.05, 0.07, 0.09}, are set to simulate the data, separately. As $\sigma$ increases, the overlaps between different groups in the simulation data also increase accordingly, which becomes more difficult to synthesize expressions accurately due to the increasing similarity between different groups. Hence, the performance of proposed method could be validated under these five conditions of $\sigma$, which could better demonstrate the robustness of the proposed method. Besides, each specific value of $\sigma$ involves five trials. Specifically, each GAN-based method under each trial is trained five times as five replicates. The expressions are synthesized in each replicate so that the synthesized expressions of different replicates are combined together for validation. In this way, the interference of instability could be reduced and the validation could be more convincing.

Table 1 demonstrates the average validation accuracies under five different $\sigma$ when identifying $\mathbf{Y}^{(2)}, \mathbf{Y}^{(3)}, \mathbf{Y}^{(4)}$, and the values in brackets are the standard deviations. Overall it can be observed that the proposed method can always achieve superior performance under most situations. Specifically, compared with PLS, the $A_1$ of the proposed method are mostly higher than the $A_1$ of PLS regression. Though sometimes the $A_1$ of PLS regression exceeds the $A_1$ of the proposed method, they are still comparable. The comparison between GRNN and the proposed SE-CTES also demonstrated similar results in terms of the $A_1$ under each $\sigma$. For the GAN-based benchmark methods, due to their inherent limitation for the characteristic-to-expression synthesis, their accuracies are not very competitive, and meanwhile it also validated the contributions of the proposed architecture in SE-CTES. Specifically, the higher accuracies of the proposed method than the CGAN demonstrate that considering three different pairs in the model training is effective. The results from the GAN-CLS and CTES are worse than the proposed method, which prove the effectiveness for assigning appropriate weights for mismatching errors and selective ensemble learning, respectively. As for the results based on $A_2$, there is no significant difference between the proposed method and other benchmark methods, and it is reasonable to obtain the similar $A_2$ of all the methods. As described in Sec. 4.1.1, when identifying group $i$, $\mathbf{Y}^{\text{test}}$ are the expressions randomly selected from group $1,..,i-1, i+1, ...,5$. Hence, $\mathbf{Y}^{\text{test}}$ are very similar to the expressions in group $1,..,i-1, i+1, ...,5$ so that they are mostly not classified into the group of $\widehat{\mathbf{Y}}^{(i)}$. Notably, the standard deviations of the proposed method are also under a relatively low level. This is because the proposed method improves the model robustness based on the selective ensemble framework. In this way, the high-quality synthesized expressions could be selected for the validation. According to the comparisons, the proposed method outperforms than other benchmark methods.

Specifically, if comparing $A_1$ according to different $\sigma$, it is shown that the accuracies of all the methods decrease as $\sigma$ increases. Such trends fit the fact that the synthesis becomes more difficult as $\sigma$ increases. In addition, for all the methods, the accuracies of identifying $\mathbf{Y}^{(4)}$ is the highest and the accuracies of identifying $\mathbf{Y}^{(2)}$ is the lowest. This may be because the difficulty to synthesize the three groups are different. Given that $x_1, x_2$ follow $N(\mu, \sigma^2)$ where $\mu = 0.2i$ for the $i$-th group, the $(x_1, x_2)$ to simulate $\mathbf{Y}^{(4)}$ is larger than the $(x_1, x_2)$ to simulate $\mathbf{Y}^{(2)}$. Then the range of the values in $\mathbf{Y}^{(4)}$ is much larger than the range of the values in $\mathbf{Y}^{(2)}$, which means the synthesis for $\mathbf{Y}^{(4)}$ is the easiest and the synthesis for $\mathbf{Y}^{(2)}$ is the hardest. No matter the comparisons are based on different $\sigma$ or different groups, the $A_1$ of the proposed method mostly remain the highest. Hence, the results demonstrate the superior performance and the relatively high robustness of the proposed method.

***Table 1*** *Accuracies and standard deviations when identifying $\mathbf{Y}^{(2)}, \mathbf{Y}^{(3)}, \mathbf{Y}^{(4)}$, separately under five different $\sigma$*

| Parameter $\sigma$ | Methods | Identify $\mathbf{Y}^{(2)}$ | | Identify $\mathbf{Y}^{(3)}$ | | Identify $\mathbf{Y}^{(4)}$ | |
|---|---|---|---|---|---|---|---|
| | | $A_1$ | $A_2$ | $A_1$ | $A_2$ | $A_1$ | $A_2$ |
| 0.01 | PLS regression | 0.300 (0.088) | 1.000 (0.000) | 0.973 (0.050) | 1.000 (0.000) | 1.000 (0.000) | 1.000 (0.000) |
| | GRNN | 0.097 (0.130) | 1.000 (0.000) | 0.960 (0.050) | 1.000 (0.000) | 1.000 (0.000) | 1.000 (0.000) |
| | CGAN | 0.332 (0.095) | 1.000 (0.000) | 0.829 (0.166) | 0.999 (0.000) | 0.995 (0.008) | 1.000 (0.000) |
| | GAN-CLS | 0.133 (0.188) | 1.000 (0.000) | 0.735 (0.091) | 1.000 (0.000) | 0.999 (0.001) | 1.000 (0.000) |
| | CTES | 0.266 (0.094) | 1.000 (0.000) | 0.704 (0.144) | 0.998 (0.003) | 1.000 (0.000) | 1.000 (0.000) |
| | **SE-CTES (Proposed)** | **0.513 (0.409)** | **1.000 (0.000)** | **0.997 (0.005)** | **1.000 (0.000)** | **0.993 (0.009)** | **1.000 (0.000)** |
| 0.03 | PLS regression | 0.660 (0.391) | 1.000 (0.000) | 0.650 (0.323) | 1.000 (0.000) | 1.000 (0.000) | 1.000 (0.000) |
| | GRNN | 0.203 (0.171) | 0.998 (0.002) | 0.663 (0.364) | 1.000 (0.000) | 0.997 (0.005) | 1.000 (0.000) |
| | CGAN | 0.441 (0.278) | 1.000 (0.000) | 0.522 (0.345) | 1.000 (0.000) | 0.935 (0.085) | 1.000 (0.000) |
| | GAN-CLS | 0.513 (0.088) | 1.000 (0.000) | 0.611 (0.071) | 1.000 (0.000) | 0.957 (0.014) | 1.000 (0.000) |
| | CTES | 0.449 (0.173) | 1.000 (0.000) | 0.653 (0.123) | 1.000 (0.000) | 0.986 (0.010) | 1.000 (0.000) |
| | **SE-CTES (Proposed)** | **0.647 (0.458)** | **1.000 (0.000)** | **0.713 (0.405)** | **1.000 (0.000)** | **1.000 (0.000)** | **1.000 (0.000)** |
| 0.05 | PLS regression | 0.879 (0.046) | 0.990 (0.008) | 0.730 (0.120) | 1.000 (0.000) | 0.780 (0.092) | 1.000 (0.000) |
| | GRNN | 0.893 (0.069) | 0.993 (0.005) | 0.873 (0.082) | 1.000 (0.000) | 0.680 (0.177) | 1.000 (0.000) |
| | CGAN | 0.389 (0.006) | 0.999 (0.001) | 0.527 (0.097) | 0.999 (0.002) | 0.646 (0.098) | 0.999 (0.001) |
| | GAN-CLS | 0.251 (0.145) | 0.998 (0.003) | 0.436 (0.131) | 1.000 (0.000) | 0.529 (0.066) | 1.000 (0.000) |
| | CTES | 0.449 (0.064) | 0.996 (0.003) | 0.615 (0.050) | 0.999 (0.001) | 0.587 (0.037) | 0.995 (0.002) |
| | **SE-CTES (Proposed)** | **0.927 (0.076)** | **0.997 (0.005)** | **0.913 (0.026)** | **1.000 (0.000)** | **0.753 (0.101)** | **0.993 (0.009)** |
| 0.07 | PLS regression | 0.497 (0.296) | 0.993 (0.005) | 0.550 (0.073) | 0.999 (0.001) | 0.963 (0.012) | 0.996 (0.002) |
| | GRNN | 0.027 (0.038) | 1.000 (0.000) | 0.633 (0.130) | 0.998 (0.002) | 0.927 (0.038) | 0.997 (0.003) |
| | CGAN | 0.219 (0.062) | 0.998 (0.002) | 0.568 (0.099) | 0.999 (0.001) | 0.836 (0.088) | 0.997 (0.002) |
| | GAN-CLS | 0.041 (0.036) | 1.000 (0.000) | 0.261 (0.192) | 0.998 (0.003) | 0.831 (0.094) | 0.998 (0.002) |
| | CTES | 0.259 (0.173) | 0.997 (0.002) | 0.532 (0.163) | 0.998 (0.002) | 0.878 (0.034) | 0.997 (0.001) |
| | **SE-CTES (Proposed)** | **0.567 (0.306)** | **0.992 (0.006)** | **0.787 (0.111)** | **0.998 (0.002)** | **0.977 (0.012)** | **0.994 (0.003)** |
| 0.09 | PLS regression | 0.177 (0.133) | 0.988 (0.007) | 0.187 (0.133) | 0.997 (0.002) | 0.347 (0.068) | 0.993 (0.001) |

|  |  |  |  |  |  |  |
|---|---|---|---|---|---|---|
| GRNN | 0.000 | 0.999 | 0.120 | 0.998 | 0.343 | 0.994 |
|  | (0.000) | (0.001) | (0.054) | (0.001) | (0.156) | (0.002) |
| CGAN | 0.022 | 1.000 | 0.113 | 0.999 | 0.249 | 0.998 |
|  | (0.023) | (0.000) | (0.052) | (0.000) | (0.082) | (0.001) |
| GAN-CLS | 0.042 | 0.999 | 0.076 | 0.999 | 0.232 | 0.998 |
|  | (0.034) | (0.001) | (0.036) | (0.000) | (0.117) | (0.002) |
| CTES | 0.113 | 0.997 | 0.128 | 0.999 | 0.246 | 0.998 |
|  | (0.079) | (0.002) | (0.101) | (0.000) | (0.026) | (0.002) |
| **SE-CTES (Proposed)** | **0.150** | **0.993** | **0.193** | **0.998** | **0.367** | **0.998** |
|  | **(0.085)** | **(0.005)** | **(0.104)** | **(0.002)** | **(0.121)** | **(0.001)** |

## 4.2 Case study for cardiovascular disease (CVD) risk prediction

Apart from the simulation study in Sec. 4.1, a real-world case study for Cardiovascular disease (CVD) risk prediction is also conducted to further demonstrate the effectiveness of the proposed SE-CTES method. In this study, the dataset is extracted from the Cohort component of the Atherosclerosis Risk in Communities (ARIC) study [35]. The selected dataset includes 10,500 samples (participants), where each sample has 1 label and 24 variables. The label represents the disease outcome, which is binary. 0 in the label means the participant was not diagnosed as CVD, especially, as fatal congenital heart defect (Fatal CHD). 10 variables describe the participants' lifestyles and unchangeable characteristics, e.g., age, BMI, alcohol consumption, which are defined as characteristics. The other 14 variables are physiological symptoms, e.g., heart rate, blood pressure, which are defined as expressions.

The existing studies like [36, 37] have shown that the risk of CVD is related to both patient characteristics and physiological symptoms (expressions). Therefore, if the relationships between the CVD risk and the variables could be quantified, CVD may be potentially prevented according to the risk prediction. However, the participants who are diagnosed with CVD are so limited that the evaluation for the risk is hard to develop due to the imbalanced data. In addition, it is also important to estimate the expected expressions if alternative lifestyles need to be evaluated. Thus, the synthesis from characteristics to expressions is also essential in this case. As shown in Figure 11, SE-CTES is applied to synthesize expressions based on characteristics. Afterwards, both characteristics and synthesized expressions are applied to predict CVD risk.

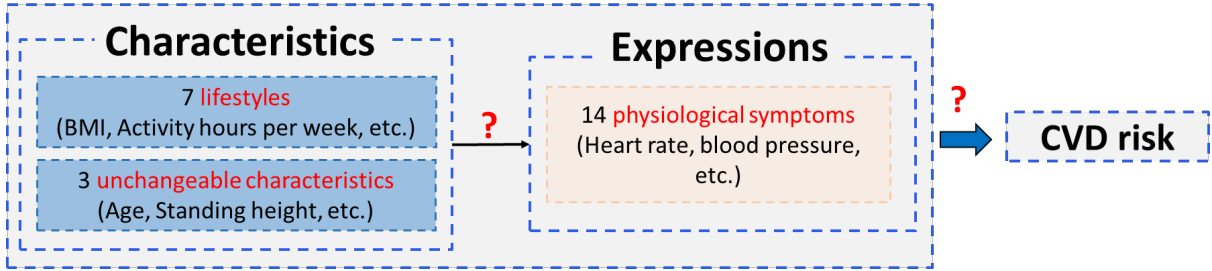

*Figure 11:* *The synthesis process of CVD case*

Notably, the continuous variables were discretized into six intervals for analysis based on quantiles. The network structure of proposed method is the same as the structure in Sec. 4.1 while the hyper-parameters, including $k$, $\beta$ and $h$, also remain the same. The same benchmark methods are selected as in Sec. 4.1, and the setups are also the same as the setup in Sec. 4.1. Since each characteristic could match one expression in this case study, the fake characteristics are randomly selected from the unmatched actual characteristics for each expression. Two different groups are extracted for the validation the performance of proposed method. The first group only contains the participants that are healthy. However, given that the unhealthy participants (i.e., diagnosed by CVD) are much less, the second group is a mixture of the participants of both healthy and unhealthy rather than only unhealthy. Each group has 3,500 participants. In this way, the proposed method could be evaluated according to both healthy and unhealthy participants.

When identifying group $i$, all the other participants except group $i$ are applied as the training set to train the proposed method. The characteristics of group $i$, $\mathbf{X}^{(i)}$, are sent to the proposed method to synthesize expressions, $\hat{\mathbf{Y}}^{(i)}$. Then $\mathbf{X}^{(i)}$ and $\hat{\mathbf{Y}}^{(i)}$ are combined as fake participants. To evaluate its effectiveness, CVD risks estimated by $\hat{\mathbf{Y}}^{(i)}$ and $\mathbf{Y}^{(i)}$ are compared. Specifically, by training a random forest classifier [34] using the actual participants, the CVD risk for the participants, $r_s$, could be predicted. Meanwhile, using the fake participants and all the other actual participants except group $i$, another random forest classifier will be trained. Then the participants in group $i$ are sent to the classifier for testing so that the CVD risk based on fake participants, $r_a$, could be predicted. The differences between $r_a$ and $r_s$ could demonstrate the quality of synthesized expressions, the smaller of $|r_a - r_s|$, the better. Hence, average $|r_a - r_s|$ could be applied as the evaluation metrics to evaluate the performance of proposed method.

Table 2 shows the average $|r_a - r_s|$ as well as the standard deviations (in brackets). Both PLS regression and GRNN have the largest differences between $r_a$ and $r_s$ no matter when identifying group 1 or group 2. As for the GAN-based benchmark approaches, they could also not achieve the differences as small as the proposed method. Thus, it is clearly shown that the proposed method remains the lowest for both the risk differences and the variations under both groups. Such results strongly prove the superior performance of the proposed method. Overall, the proposed SE-CTES method can stably synthesize accurate expressions under the given characteristics.

*Table 2: Average $|r_a - r_s|$ and standard deviations of identifying group 1,2 separately*

| Method | Identify group 1 (healthy) | Identify group 2 (mixed) |
|---|---|---|
| PLS regression | 0.2381 (0.0807) | 0.1412 (0.0775) |
| GRNN | 0.2394 (0.0855) | 0.1483 (0.0831) |
| CGAN | 0.1918 (0.0735) | 0.1160 (0.0636) |
| GAN-CLS | 0.2061 (0.0813) | 0.1238 (0.0696) |
| CTES | 0.1943 (0.0919) | 0.1120 (0.0715) |
| **SE-CTES (Proposed)** | **0.1765 (0.0713)** | **0.1078 (0.0636)** |

## 5 Extended simulation study for scalar to matrix synthesis

The case studies in Sec. 4 demonstrate the superior performance of proposed method, where both the characteristics and expression are multivariate data. To further explore the capability of the proposed method when the dimension of expressions become much higher, for example, the matrix-based expressions, an extended simulation study for scalar-based characteristic to matrix-based expression synthesis, i.e., mapping from scalar-based **x** to matrix-based **y**, is developed. Initially, the Gaussian process (GP) is applied to simulate expressions of five categories, $\mathbf{Y}^{(1)}, \mathbf{Y}^{(2)}, \mathbf{Y}^{(3)}, \mathbf{Y}^{(4)}, \mathbf{Y}^{(5)}$, as shown in Eq. (14) where $\kappa$ is applied as the radial basis function (RBF) kernel. 128 images with dimension $64 \times 64$ are simulated in each category. In this way, though the parameter applied to simulate expressions in one category is the same, the expressions in one category are different so that the stochastic patterns are involved.

$$\mathbf{Y}^{(i)} = \begin{bmatrix} \mathbf{y}_{i,1} \\ \vdots \\ \mathbf{y}_{i,128} \end{bmatrix}, \quad \mathbf{y}_{i,j} \sim GP(0, \kappa),$$

$$\kappa\left(\mathbf{y}_{i,j_m}, \mathbf{y}_{i,j_n}\right) = \exp\left(-\frac{1}{2l_i}(\mathbf{y}_{i,j_m}{}^2 + \mathbf{y}_{i,j_n}{}^2)\right), \quad l_i = i, \quad (14)$$

$$i = 1,2,3,4,5, \quad j = 1,2,\dots,128, \quad m, n = 1,2,\dots,64$$

The examples from the five categories, $\mathbf{y}_1, \mathbf{y}_2, \mathbf{y}_3, \mathbf{y}_4, \mathbf{y}_5$, are demonstrated as shown in Figure 12. As the parameter $l_i$ increases, the images become smoother.

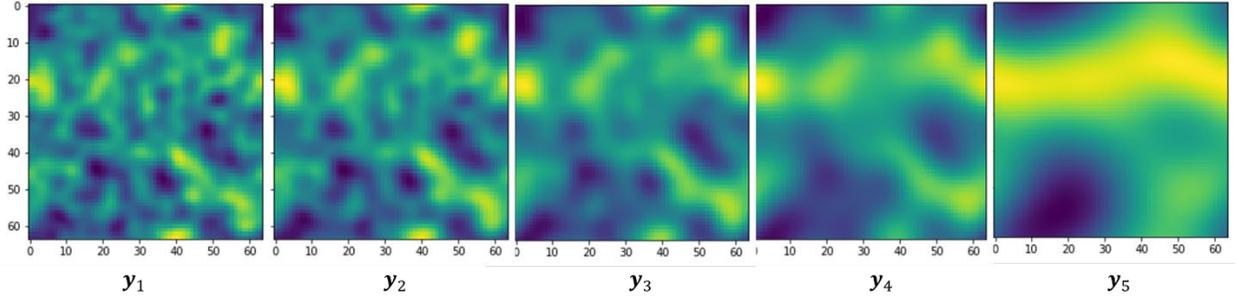

*Figure 12: Examples from 5 different categories of simulated expressions*

According to Eq. (14), one $l_i$ simulates 128 images. In order to increase the diversity of characteristics, the extensions of $l_i$ is developed based on normal distributions for synthesis. Hence, five groups of characteristics, $\mathbf{X}^{(1)}, \mathbf{X}^{(2)}, \mathbf{X}^{(3)}, \mathbf{X}^{(4)}, \mathbf{X}^{(5)}$, corresponding to $\mathbf{Y}^{(1)}, \mathbf{Y}^{(2)}, \mathbf{Y}^{(3)}, \mathbf{Y}^{(4)}, \mathbf{Y}^{(5)}$, are also simulated from $l_i$ as Eq. (15).

$$\mathbf{X}^{(i)} = \begin{bmatrix} \mathbf{x}_{i,1}^T \\ \vdots \\ \mathbf{x}_{i,128}^T \end{bmatrix} \Leftarrow \mathbf{x}_{i,j} = \begin{bmatrix} v_1 \\ \vdots \\ v_{64} \end{bmatrix}^T \Leftarrow v_q \sim N\left(20l_i + \frac{q}{10}, 1\right) \quad (15)$$

$$i = 1,2,3,4,5, \quad j = 1,2,\dots,128, \quad q = 1,2,\dots,64$$

Similar as Sec. 4.1, $\mathbf{Y}^{(2)}, \mathbf{Y}^{(3)}, \mathbf{Y}^{(4)}$ are separately identified for model validation, which also follows the model validation framework presented in Sec. 3.4. A CNN classifier with four convolutional layers is applied for both inverse validation and model validation. All the benchmark approaches described in Sec.

4.1 are also applied in this simulation study. Especially for the PLS, a CNN autoencoder is first trained to transform the expressions to feature vectors. Then feature vectors are applied in the PLS as responses. In addition, the neural network structures of GRNN in this simulation study apply CNN rather than MLP to better fit the expressions. As for the GAN-based benchmark methods, given that the expressions are matrix, $\varphi$ and $\phi$ are also applied with four convolutional layers, separately. Besides, the network structures of $h_1$ and $h_2$ also follow the generator and discriminator of [9], separately. Furthermore, in order to make the synthesized expressions smoother, a low-pass filter kernel of size $(3,3)$ is applied on the synthesized expressions. All the methods are applied under three trials, and each GAN-based method under each trial is also trained five times as five replicates. The same validation metrics in Sec. 4.1, $A_1$ and $A_2$, are also applied to validate the performance of proposed method.

Thereby, Table 3 demonstrates the average validation accuracies of five trials and the values in brackets are the standard deviations. For $A_1$, the proposed method remains the highest no matter when identifying which group. The standard deviations of the proposed method are also under a relatively low level. As for $A_2$ and its standard deviations, though the proposed method is not the highest, they are still acceptable. Thus, the results demonstrate the outperformance of proposed method than other benchmark methods under the case with matrix-based expressions.

**Table 3** *Accuracies and standard deviations when identifying $\mathbf{Y}^{(2)}, \mathbf{Y}^{(3)}, \mathbf{Y}^{(4)}$, separately*

| Method | Identify $\mathbf{Y}^{(2)}$ | | Identify $\mathbf{Y}^{(3)}$ | | Identify $\mathbf{Y}^{(4)}$ | |
|---|---|---|---|---|---|---|
| | $A_1$ | $A_2$ | $A_1$ | $A_2$ | $A_1$ | $A_2$ |
| PLS regression | 0.010 (0.008) | 0.957 (0.003) | 0.118 (0.071) | 0.935 (0.013) | 0.676 (0.136) | 0.937 (0.012) |
| GRNN | 0.331 (0.496) | 0.888 (0.112) | 0.129 (0.114) | 0.981 (0.016) | 0.341 (0.496) | 0.888 (0.112) |
| CGAN | 0.222 (0.199) | 0.903 (0.002) | 0.000 (0.000) | 0.908 (0.002) | 0.248 (0.233) | 0.919 (0.065) |
| GAN-CLS | 0.047 (0.032) | 0.973 (0.022) | 0.074 (0.043) | 0.943 (0.055) | 0.301 (0.270) | 0.944 (0.016) |
| CTES | 0.351 (0.008) | 0.952 (0.008) | 0.176 (0.168) | 0.929 (0.015) | 0.383 (0.227) | 0.925 (0.015) |
| **SE-CTES (Proposed)** | **0.445 (0.101)** | **0.957 (0.028)** | **0.352 (0.337)** | **0.931 (0.009)** | **0.726 (0.086)** | **0.947 (0.037)** |

# 6 Conclusions

In this paper, a novel GAN-based selective ensemble characteristic-to-expression synthesis (SE-CTES) is proposed to synthesize the medical expressions based on given characteristics for healthcare analytics. Compared with the existing synthesis models, the proposed CTES algorithm involves three contributions: (1) GAN-based neural networks are incorporated to quantify the mapping from relatively low dimensional characteristics to high dimensional expressions; (2) the suitable weights of two mismatching errors are considered in model training and optimized to reduce the learning bias; (3) a novel selective ensemble learning framework is proposed to improve the synthesis stability and further reduce the prediction bias by the proposed inverse validation framework.

The superior performance of CTES is demonstrated by extensive numerical simulation studies and a real-world case study. Based on the trials under different values of parameter $\sigma$, the first simulation study strongly proves the effectiveness and the relatively high robustness of the proposed method. Then a case study for CVD disease is developed and the results show that the proposed method outperforms than all the other benchmark methods. Afterwards, an extended simulation study for scalar to matrix synthesis is also demonstrated, which also shows the potential feasibility of proposed method for synthesizing matrix-based expressions. Therefore, the proposed SE-CTES method is very promising for the synthesis of expressions under given characteristics.